\documentclass[10pt, conference]{IEEEtran}
\IEEEoverridecommandlockouts
\usepackage[switch]{lineno}
\usepackage{multicol,lipsum}    
\usepackage{cite}
\usepackage{amsmath,amssymb,amsfonts}
\usepackage{graphicx}
\usepackage{textcomp}
\usepackage{xcolor}
\def\BibTeX{{\rm B\kern-.05em{\sc i\kern-.025em b}\kern-.08em
    T\kern-.1667em\lower.7ex\hbox{E}\kern-.125emX}}
\usepackage{blindtext}
\usepackage{graphicx}
\usepackage{comment}
\usepackage{tabularx}
\usepackage{placeins}
\usepackage{makecell}

\usepackage{hyperref}
\hypersetup{
    colorlinks=true,
    linkcolor=blue,
    filecolor=cyan,      
    urlcolor=magenta
}
\usepackage{xspace}
\usepackage{subfigure} 
\usepackage{graphics} 
\usepackage{epsfig} 
\usepackage{multirow}
\usepackage{bbm}
\usepackage{caption}
\usepackage{tabularx}

\usepackage{bm}
\usepackage{bbold}
\newcommand{\mbf}[1]{\mathbf{#1}}

\usepackage{algpseudocode}

\let\oldReturn\Return
\renewcommand{\Return}{\State\oldReturn}
\usepackage{algorithm}
\usepackage{algorithmicx}
\usepackage{graphicx}
\usepackage{textcomp}
\usepackage{xcolor}
\def\BibTeX{{\rm B\kern-.05em{\sc i\kern-.025em b}\kern-.08em
    T\kern-.1667em\lower.7ex\hbox{E}\kern-.125emX}}
    
\newcommand{\eatme}[1]{ }

\begin{document}

\title{SparsePipe: Parallel Deep Learning for 3D Point Clouds}

\author{\IEEEauthorblockN{Keke Zhai\IEEEauthorrefmark{2},
Pan He\IEEEauthorrefmark{2}, 
Tania Banerjee, Anand Rangarajan, and Sanjay Ranka} 
Department of Computer and Information Science and Engineering, University of Florida \\
\IEEEauthorrefmark{2} denotes equal contributions \\
\{zhaikeke;pan.he;tmishra;anandr;sranka\}@ufl.edu}

\maketitle

\begin{abstract}
We propose SparsePipe, an efficient and asynchronous parallelism approach for handling 3D point clouds with multi-GPU training. SparsePipe is built
to support 3D sparse data such as point clouds. It achieves this by adopting generalized convolutions with sparse tensor representation to build expressive high-dimensional convolutional neural networks. Compared to dense solutions, the new models can efficiently process irregular point clouds without densely sliding over the entire space, significantly reducing the memory requirements and allowing higher resolutions of the underlying 3D volumes for better performance.

SparsePipe exploits intra-batch parallelism that partitions input data into multiple processors and further improves the training throughput with inter-batch pipelining to overlap communication and computing. Besides, it suitably partitions the model when the GPUs are heterogeneous such that the computing is load-balanced with reduced communication overhead. 

Using experimental results on an eight-GPU platform, we show that SparsePipe can parallelize effectively and obtain better performance on current point cloud benchmarks for both training and inference, compared to its dense solutions. %
\end{abstract}


\section{Introduction}\label{sec:intro}
Point clouds are captured by 3D scanners, light detection and ranging (LiDAR), structure-from-motion (SFM) techniques, and recently available 3D sensors, such as Kinect and Xtion \cite{paolanti2020multidisciplinary}. Point clouds are used widely in various applications such as virtual reality, 3D gaming, and digital preservation. There is an increasing interest in applying deep learning approaches to point cloud data \cite{graham20183d,qi2017pointnet,qi2017pointnet++}. Point clouds usually have a sparse point density, especially when compared to the continuous actual surface. In many of the current approaches, a preprocessing step (e.g,. dense image matching \cite{remondino2013dense}) is thereby applied that transforms the point clouds into dense tensors. Subsequently, the dense tensors are processed using deep learning models where the core operations are a set of the regular dense convolutions. Another approach to applying convolution to point clouds quantizes the entire space into 3D voxels and then densely convolves them 
in a sliding window fashion \cite{dai2017scannet}.

Unfortunately, converting the point clouds into 3D dense tensors results in a large memory footprint and additional computations. Further, using data-parallel pipelining approaches to speed up the training processes would replicate the dense model to each processor. Overall, this results in high memory requirements for the weight matrix and corresponding synchronization costs to collect all gradients, update the weight matrix, and redistribute to each processor. In addition to the above, there are additional practical challenges. Due to the unstructured nature of point clouds, constructing \textit{convolutions} for point clouds requires expensive nearest neighbor search such as KD-Tree or Ball Query. This makes 
it nontrivial to integrate point cloud models into existing dense computation frameworks, which partially explains why most point cloud models (e.g., dense 3D convolution~\cite{dai2017scannet}, PointNet-variants\cite{qi2017pointnet,qi2017pointnet++}) are shallow models with 
few layers. 

In this paper, we develop SparsePipe, which addresses many of the above limitations. It supports storage of data using a sparse tensor representation and generalized convolutions for handling 3D point clouds. Using sparse models is both memory and computation efficient and enables us to solve larger problems with higher voxelization resolutions and deeper models. Additionally, it provides fast parallelization of these representations on a heterogeneous cluster of GPU processors as compared to naive data parallelism.

To the best of our knowledge, this is the first work that addresses sparse computation frameworks on multiple GPUs with pipeline model parallelism.

The \textit{SparsePipe} has the following key contributions:
\begin{enumerate}

\item  It uses sparse tensor representation for processing 3D point clouds that has a relatively small memory footprint as compared to dense approaches.

\item It integrates model parallelism with data parallelism and processes mini-batches in a pipelined fashion. The parallelization algorithms in SparsePipe are based on PipeDream \cite{narayanan2019pipedream} and SpecTrain \cite{chen2018efficient}, which are limited for dense models.
\item It incorporates a load-balancing step that is aware of the underlying platform and exploits differential GPU characteristics by suitably partitioning the overall pipeline.  
\end{enumerate}
Overall, this results in both effective utilization of the underlying parallelization and computation resources while supporting much larger 3D datasets due to a smaller footprint. It achieves higher accuracy compared to dense baselines for shape classification. Compared to data-parallel training, SparsePipe is faster while maintaining high accuracy. 
 
\section{Background}\label{sec:background}
 
\subsection{DNN Model Training}
A deep neural network (DNN) model consists of multiple consecutive layers. The goal of training the DNN model is to find its optimal set of parameters (weights) $w$ that can minimize the sum of the objective function for training samples with labels. This is  
commonly accomplished using \textit{stochastic gradient descent }(SGD) \cite{robbins1951stochastic}. This approach computes the weight updates, i.e., \textit{gradients}, on a given mini-batch (a subset of training samples) and updates the weight $w$. The size of the mini-batch is chosen based on convergence and processing requirements. The training is decomposed into \textit{forward} and \textit{backward}. The \textit{forward} computation makes predictions of given samples, and each layer computes its activations to be fed into the next layer given its current input and layer parameters. The \textit{backward} computation computes the loss in the end layer and backpropagates it through all layers. The gradients for each layer (including its inputs and weights) are estimated using gradients from upper layers and previously computed layer activations. The SGD optimizer then updates the model parameters based on these gradients.

Parallel computing algorithms that have been successfully applied to speed up the DNN model training, which mainly contains two broad classes: intra- and inter-batch parallelism. The advantages and limitations of these approaches are described in the following subsections.

\subsection{Intra-batch Parallelism}

The intra-batch parallelism splits a single training iteration across processors. Two popular types of intra-batch processing are widely adopted in distributed deep learning frameworks: data parallelism (DP) and model parallelism (MP).

\textbf{Data Parallelism}. DP is the most common approach, in which a model is replicated and distributed to multiple processors such that each model handles a  subset of the input dataset. The forward- and backward-computations are performed at each processor. Weight updates are aggregated by communicating and synchronizing between processors to obtain a final weight update. The amount of data communicated between processors is therefore proportional to the size of the model. Data parallelism is the most popular and practical way of performing distributed parallel training due to its flexibility and wide support across popular deep learning frameworks such as PyTorch \cite{paszke2017automatic}, TensorFlow \cite{abadi2016tensorflow}, and Caffe \cite{jia2014caffe}. For large models, the communication overhead can be high because the weights are replicated across processors and they have to be updated frequently. This overhead increases as the number of processors increases. Even with the use of high-performance communication libraries such as the NVIDIA Collective Communications Library (NCCL), communication overhead can be as large as 
85\% of training time for the convolutional neural network model VGG16 \cite{simonyan2014very}. Other optimization approaches have been used for DP: asynchronous parallelism for hardware efficiency \cite{chen2016revisiting}, gradient quantization for reducing sizes of data to be communicated between processors \cite{seide2016cntk}, and specialized network hardware for reducing communication overheads. These methods are complementary to DP. Recent approaches follow layer-wise adaptive rate scaling (LARS) \cite{goyal2017accurate} for training models effectively with large mini-batches, which reduces the communication overhead with fewer parameters exchanged.

\begin{figure}[hbt!]
    \centering
    \includegraphics[width=0.5\textwidth]{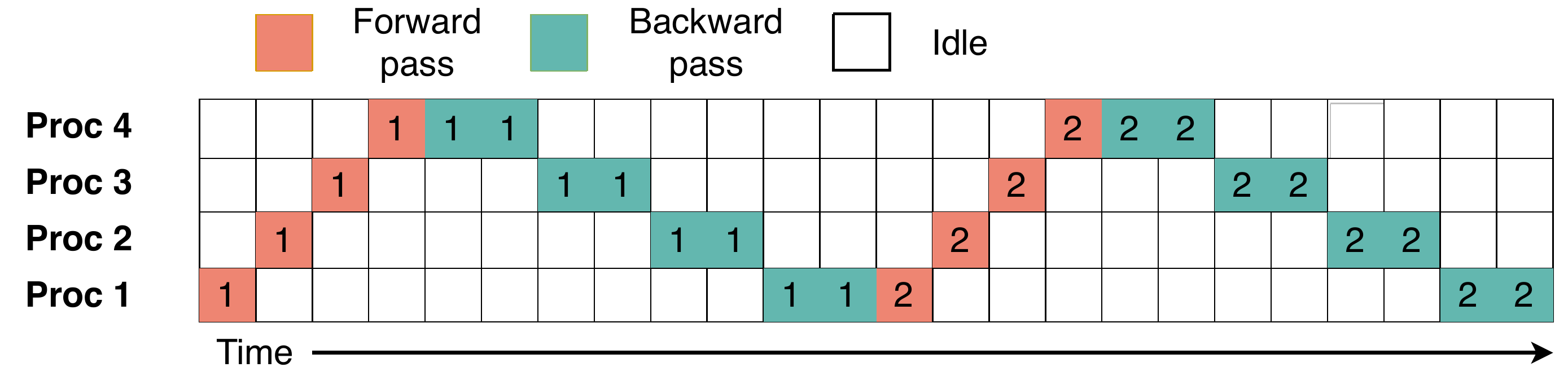}
    \caption{Naive pipeline}
    \label{fig:naive_mp}
\end{figure}

\begin{figure}[hbt!]
    \centering
    \includegraphics[width=0.5\textwidth]{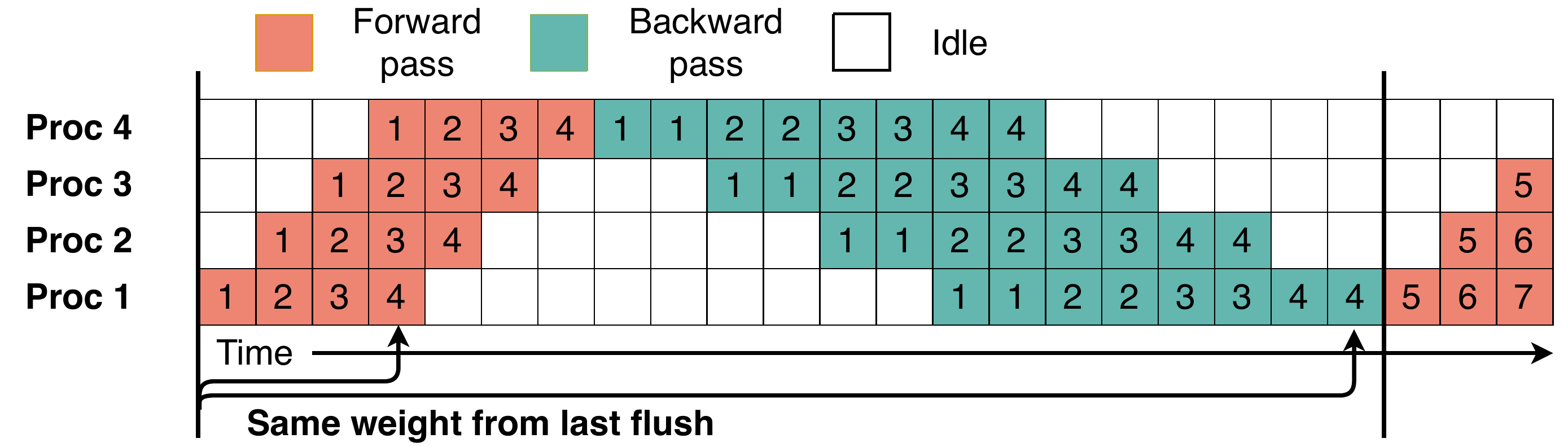}
    \caption{GPipe \cite{huang2019gpipe}}
    \label{fig:gpipe}
\end{figure}

\begin{figure}[hbt!]
    \centering
    \includegraphics[width=0.5\textwidth]{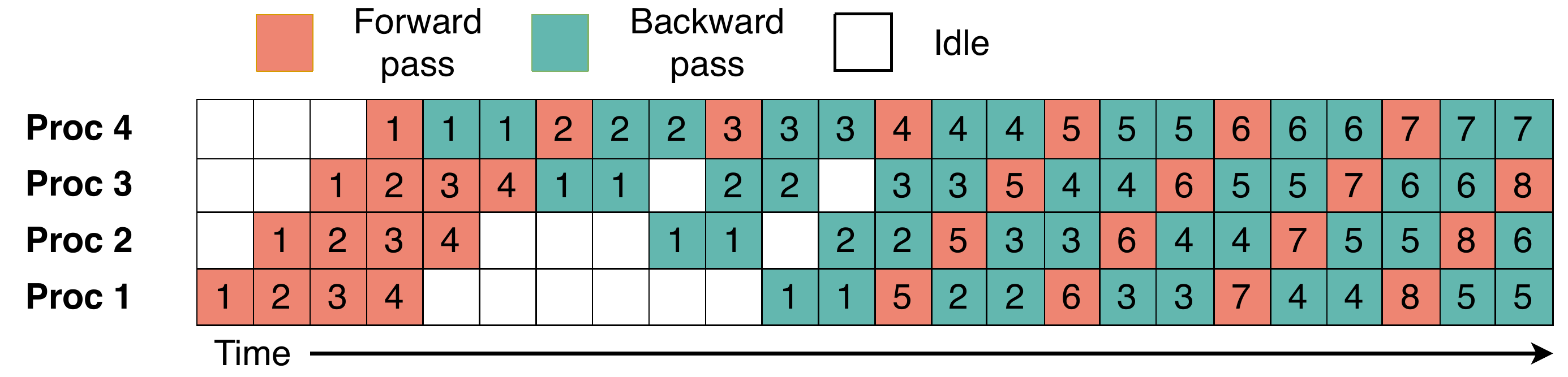}
    \caption{Pipedream \cite{narayanan2019pipedream}}
    \label{fig:pipedream}
\end{figure}

\textbf{Model Parallelism}. MP partitions a model among processors so that each processor only updates a subset of the model weights. Compared to data parallelism, model parallelism reduces data communication by sharing intermediate outputs (and the corresponding gradients). However, vanilla MP is rarely adopted in practical  due to several major limitations. As illustrated in Fig. \ref{fig:naive_mp}, it suffers from under-utilization of GPU accelerators. The data dependency (including activations and gradients) between processors will make only one processor active while stalling others at the same time. Moreover, MP highly relies on a proper model partitioning to have similar throughput for all the sub-models to avoid any GPU under-utilization, and obtaining an optimal partitioning is non-trivial. Heuristic partitions by programmers usually obtain point solutions that are far from the optimal ones.

\textbf{Hybrid Intra-batch Parallelism}. A natural extension is to partition a model while regarding both DP and MP. FlexFlow \cite{jia2019beyond} introduced a novel execution similar to finding a fast parallelization strategy to split one iteration. Krizhevsky's OWT ("one weird trick") \cite{krizhevsky2014one} explored the AlexNet model and conducted data parallelism for convolutional layers while choosing not to replicate fully connected layers with a large number of model parameters. We refer the interested reader to \cite{ben2019demystifying} for a detailed review.

\subsection{Inter-batch Parallelism} \label{sec:background_interBatch}
The inter-batch parallelism tries to parallelize the computation across multiple batches. Most of them use the pipeline parallel (PP) paradigm to address the limitations of MP and DP approaches \cite{chen2012pipelined,huang2019gpipe,narayanan2019pipedream}.  Instead of just one input, multiple inputs are injected into a computation pipeline. This ensures a better schedule that reduces waiting time and improves the utilization of computing resources (Figs. \ref{fig:gpipe} and~\ref{fig:pipedream}). 

To address the low GPU-utilization in naive model parallelism,
GPipe \cite{huang2019gpipe} starts by splitting mini-batch training samples into smaller \textit{micro-batches}, which allows a finer-grained training unit. GPipe trains these training units in a pipeline fashion, which, to some extent, allows a concurrent training on multiple GPUs, thus significantly improving the GPU utilization. GPipe is considered to be a synchronous parallelism where the micro-batches are processed sequentially, which inevitably causes some bubble overheads. To maintain the model accuracy, a single version of model weights is maintained, and the model's weights are periodically updated (\textit{flushed}) during the pipeline training.

Asynchronous parallelism \cite{gaunt2017ampnet} techniques are used to improve the GPU utilization with asynchronous weight update immediately after sufficient gradients are accumulated. It allows each processor to proceed with the next input minibatch before receiving the gradients from the previous minibatch thus overcoming
 the low device-utilization of the naive model parallelism, as shown in~\cite{narayanan2019pipedream}. However, applying this to the naive pipelined system will encounter the problem widely referred to as the weight inconsistency, due to the fact that each processor sees two mismatched model parameters of the same mini-batch samples during its \textit{forward} and \textit{backward} passes. This discrepancy in weights can prevent model convergence. To address this issue, PipeDream \cite{narayanan2019pipedream} proposed the \textit{weight stashing} technique to mitigate the weight discrepancy issue and ensure that the same version of model parameters is used for each mini-batch sampling \textit{within} a stage. The \textit{vertical sync} is provided as well for solving inconsistency of model parameters \textit{across} stages. Still, PipeDream suffers from the weight staleness issue where different versions of weights are used across all training iterations, which is partly addressed in SpecTrain \cite{chen2018efficient} via weight prediction. The weight prediction is based on the observation that gradients are smoothed in a momentum-based optimizer.  Therefore, future weights can be potentially predicted in early pipeline stages to replace the stale weights presented in current PipeDream.

\subsection{3D Point Cloud Processing}

\textbf{Models with 3D Convolutions.} One conventional approach for processing 3D point clouds is to adopt volumetric representation. The early work uses a rectangular grid or dense representation that represents the space either as \textbf{0/1} or the signed distance function (SDF), followed by 3D convolutional neural networks (3D CNNs). Huang and You~\cite{huang2016point} introduced a labeling scheme using a simple 3D CNN network for point cloud processing. Given a big point cloud and a center point, they set up a cubic bounding box with a defined radius around the center reference point. The model divides the cube into a grid of cells, which are further transformed into voxelized occupancy grids. The label for each voxel is inferred using a voting scheme.

All major public neural network frameworks can support 3D CNN operators based on this straightforward representation. However, they suffer from expensive computation and high memory consumption, which limits their use to processing of point clouds at a very coarse resolution (typically on the order of $32 \times 32 \times 32$). The inefficient dense sliding window techniques for 3D CNNs further limit the receptive field of a model because only shallow models with fewer layers are applicable.
To mitigate these issues, the octree representation is incorporated into the volumetric CNNs \cite{riegler2017octnet}. A potentially more viable approach is based on the use of pseudo-convolutional neural networks where the key idea is to define convolutions using continuous kernels, assuming a continuous space for point clouds. The major limitation is an expensive nearest neighbor search for the kernels, even with an efficient implementation of KD-Tree. Fortunately, sparse 3D CNNs have received more attention because of only non-empty locations with a small percentage of the entire space needing to be processed. Several frameworks (such as SparseConvNet \cite{graham20183d} and MinkowskiEngine \cite{choy20194d}) can compute the sparse CNNs based on the efficient indexing structure. Other alternative solutions, e.g., sparse blocks network (SBNet) \cite{ren2018sbnet}, are available as well.

\textbf{Models without 3D Convolutions.}
Deep learning models have been proposed to process point clouds without 3D convolutions. In \cite{pan2018convolutional}, the authors conducted 2D convolutions directly on the surfaces for segmentation. PointNet \cite{qi2017pointnet} directly treats a set of point coordinates as features and applies a multilayer perceptron, followed by permutation-invariant operators (e.g., the global max-pooling layer) to obtain the global features for the classifier. The major limitation of PointNet is that it does not capture local structure, which an important facet  of the success of convolutional architectures. PointNet++ \cite{qi2017pointnet++} introduced a hierarchical neural network that can partition points into overlapped local regions and extract local features accordingly via a mini-PointNet. Later, many model variants further improved the performance with more advanced abstract layers for extracting local structure.

\section{SparsePipe}\label{sec:method}

In this section, we introduce the pipeline parallel framework SparsePipe.  This leverages state-of-the-art parallel computing frameworks presented in PipeDream \cite{narayanan2019pipedream} that have been developed for dense data processing (e.g., 2D images). We handle sparse 3D data  (in particular point clouds) by building expressive high-dimensional convolutional neural networks.  The details of the integration with sparse tensors and the generalized convolutions are presented. Finally, we present a  model partitioning algorithm that automatically partitions the model layers among different types of processors, with the goal to keep the load as balanced as possible.

\subsection{Pipeline Model Parallel} \label{subsec:pipeline}
In Section \ref{sec:background_interBatch}, we have briefly introduced the pipeline parallelism and its two main applications: GPipe \cite{huang2019gpipe} and PipeDream \cite{narayanan2019pipedream}. In this section, the asynchronous pipeline model parallelism utilized in PipeDream is introduced in a more detailed way since SparsePipe leverages this pipeline parallel approach and presents several improvements.

Pipeline parallelism of the underlying model partitions the DNN models into several stages, with each stage containing a continuous chunk of model layers. In PipeDream, each stage can be assigned to one GPU or multiple GPUs. Fig.~\ref{fig:pipedream} gives an example of the data computation among four GPUs with PipeDream, where each GPU takes charge of one stage. The $x$-axis represents time, where we assume backward computation takes twice the amount of time as the forward computation. In the warm-up stage, the processor in stage $1$ starts the training by forwarding multiple mini-batches in a row to ensure enough workloads in the pipeline. In the steady state, each GPU performs one forward computation followed by one backward computation. The weight version used for computing the forward pass and backward pass of one mini-batch is different if applying naive pipeline parallelism. For example, in stage $1$, the forward computation of mini-batch $5$ is performed right after the weights are updated by mini-batch $1$, whereas the backward computation of mini-batch $5$ is conducted with the weights updated by mini-batch $2$, $3$, and $4$. \eatme{Several challenges need to be addressed. (1) The discrepancy of weights between the forward pass and backward pass of the same mini-batch samples will reduce the model accuracy or even cause model divergence. (2) The inconsistency of model weights across stages further aggravates the problem. For example, mini-batch $5$ uses weight updates from mini-batch $1$ for processor $1$ and from $2$ for processor $2$, and so on. }PipeDream uses a technique called \textit{weight stashing}, which stores the old weights and applies them during the backward computation to the mini-batches using the same weight as in the forward pass, which mitigates the weight discrepancy and guarantees the same weight version for each mini-batch in one stage.

\begin{figure}[bt!]
    \centering
    \includegraphics[width=0.5\textwidth]{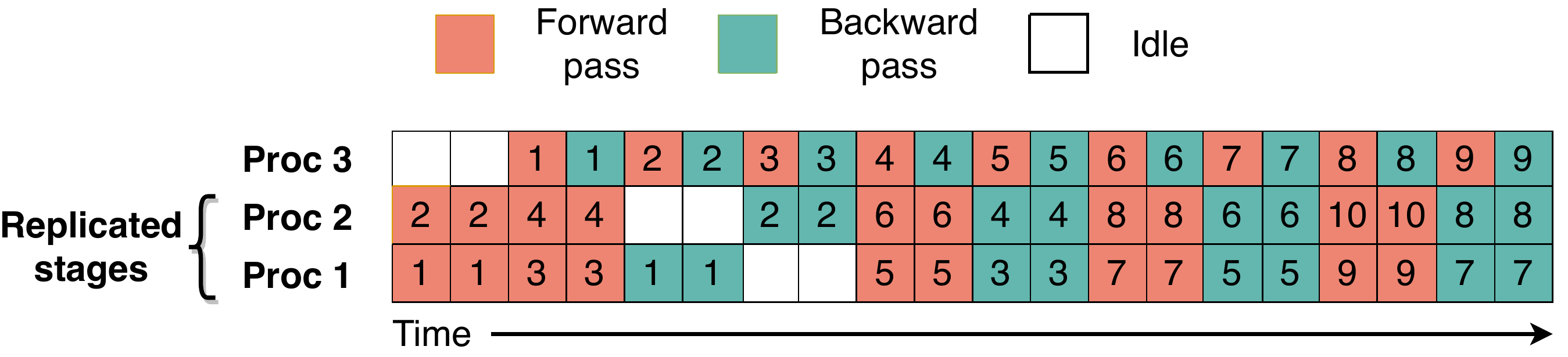}
    \caption{Hybrid pipeline model parallelism with 3 processors and 2 stages. Stage 1 takes twice of the time units than stage 2 in the forward or backward pass. To sustain roughly the same throughput of stage 2 (on proc 3), stage 1 is replicated on two processors (proc 1 and 2).}
    \label{fig:replicated_pipedream}
    \vspace{-0.5cm}
\end{figure}

Additionally, PipeDream provided a hybrid pipeline model parallelism that combines the data parallel and pipeline model parallel (Fig. \ref{fig:replicated_pipedream}). The DNN model is partitioned into 2 stages, with the first stage replicated among two GPUs. This part is done by integrating PyTorch's Distributed Data Parallel library \cite{paszke2017automatic}. A deterministic round-robin strategy is used to distribute the intermediate results from the previous duplicated stage to the next stage.  This is calculated based on the mini-batch ID and the number of replicas in the current and next stage. Although simple, it guarantees every mini-batch is calculated in the same way during the forward and backward passes, which is necessary for the saved parameters and intermediate results applied during the backward calculation.

SparsePipe leverages the asynchronous pipeline model parallelism shown above. It also incorporates the naive data parallelism. This can be viewed as the DNN model partitioned into one stage which is replicated among multiple GPUs.

\subsection{Extension to Sparse Data}\label{sec:method_sparse}
Prior work on parallelization of deep networks \cite{huang2019gpipe,narayanan2019pipedream} has shown its benefits by speeding up training for dense tensor processing (e.g., 2D images) where the core operations are the regular dense convolutions with features represented using dense formats. Dense representations are not efficient for 3D point cloud data because much of the spatial volume is empty and has no features. Hence, these approaches are not very useful.  We now describe our approach that leverages sparse tensor-based generalized convolutions.

\begin{figure}[bt!]
    \centering
    \includegraphics[width=0.35\textwidth]{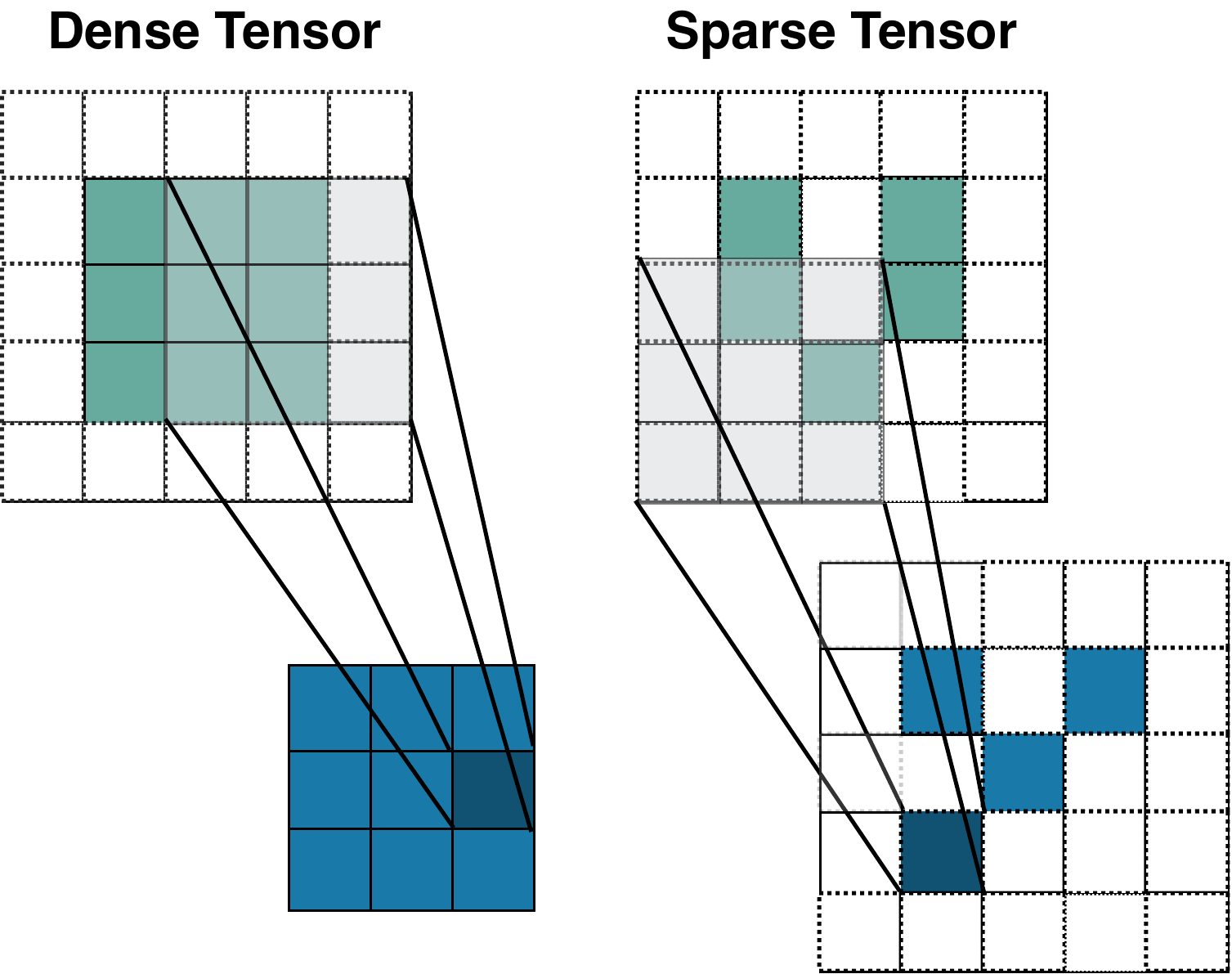}
    \caption{Visualization of the convolution operators conducted on dense  and sparse tensors. Blue and green denote the input and out feature maps, respectively. Gray indicates the convolutional kernel. It  will densely slide over the entire space. On a sparse tensor, the convolution is instead only conducted on a few specified locations.}
    \label{fig:sparse_conv}
    \vspace{-0.5cm}
\end{figure}

\begin{figure*}[hbt!]
    \centering
    \includegraphics[width=\textwidth]{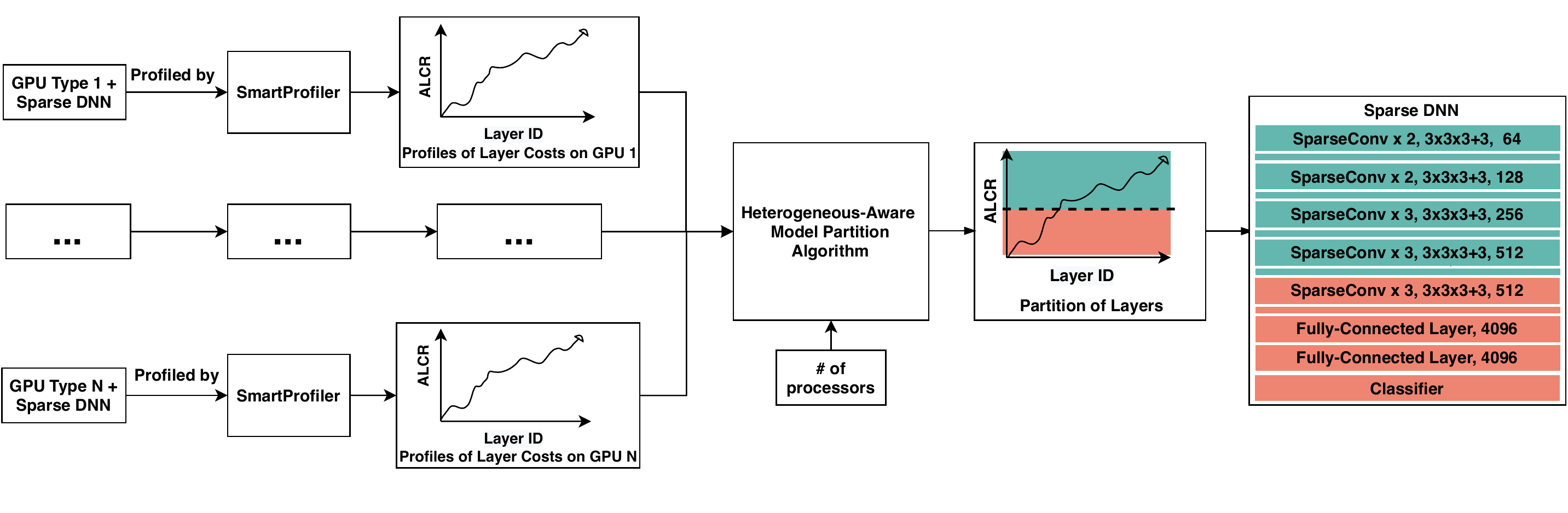}
    \vspace{-0.8cm}
    \caption{An overview of the heterogeneous-aware pipeline model partition algorithm for sparse-tensor-based computation. Given an input sparse DNN, we profile the computation time, communication time, and parameter sizes for layers, from which we compute the accumulated layer costs ratio (ALCR), with increasing layer IDs. Our method measures across multiple machines with GPU processors that are heterogeneous. We then aggregate all profiles and use a dynamic programming algorithm to obtain feasible model partition that divides the whole model into sub-models and distributes to processors. For simplicity, we omit drawings of all intermediate layers (e.g., batch normalization, pooling, or activation layers) between SparseConv blocks of the Sparse DNN.}
    \label{fig:hpc}
\vspace{-0.5cm}
\end{figure*}

Sparse tensors \cite{choy20194d} (unlike their dense counterparts) only save the non-empty part of the space thus resulting in a compact representation. We represent data with the sparse tensor, which represents as follows:
\begin{equation}
    C = \begin{bmatrix}
        b_1 & x_1 & y_1 & z_1\\
        b_2 & x_2 & y_2 & z_2\\
            & \vdots  & \vdots &  \\
        b_N & x_N & y_N & z_N\\
        \end{bmatrix}, 
    F = \begin{bmatrix}
        \mathbf{f}_1^T \\
        \mathbf{f}_2^T\\
        \vdots\\
        \mathbf{f}_N^T
        \end{bmatrix}
\end{equation}
where $C$ of size $(N, D_c + 1)$ represents the coordinates matrix and $F$ of size $(N, D_f)$ denotes its corresponding feature matrix. $N$ is the total number of points in batches; $D_c$ and $D_f$ are the coordinate and feature dimensions. Each row in coordinate matrix $C$ stands for a point location, with $b_i$ being the batch indices of $i$ points that are used to distinguish points at the same location in different batches, and $(x_i, y_i, z_i)$ is the point coordinates. The feature matrix $F$ contains a set of features with $\mathbf{f}_j$ at $j$-th row being the feature vector located at $(b_j, x_j, y_j, z_j)$ in $C$. This sparse tensor representation can be extended to 4D or higher dimensions.

The sparse tensor convolution generalizes and extends dense convolution computation \cite{paszke2017automatic,choy20194d}. The visualization of the dense tensor and sparse tensor convolution is presented in Fig.~\ref{fig:sparse_conv}. Denote by $x^{\text{in}}_\mbf{u} \in \mathbb{R}^{N^\text{in}}$ the $N^\text{in}$-dimensional input feature vector at the point $\mbf{u} \in \mathbb{R}^D$ (a D-dimensional coordinate), and $\mathbf{W} \in \mathbb{R}^{K^D
\times N^\text{out} \times N^\text{in}}$ the convolutional kernel weights, which is split into spatial weights with $K^D$ matrices of size $N^\text{out} \times N^\text{in}$ as $W_\mbf{i}$. The conventional dense convolution  is defined as: 
\begin{equation}
\mathbf{x}^{\text{out}}_\mbf{u} = \sum_{\mbf{i} \in \mathcal{V}^D(K)} \mathbf{W}_\mbf{i} \mathbf{x}^{\text{in}}_{\mbf{u} + \mbf{i}} \text{ for } \mbf{u} \in \mathbb{Z}^D,
\label{eq:dense_convolution} 
\end{equation} 
where $\mathcal{V}^D(K)$ is the list of offsets centered at the origin. e.g., $\mathcal{V}^1(3)=\{-1, 0, 1\}$. $\mathbb{Z}^D$ indicates that the convolution is conducted in the entire space by densely sliding over all positions. However, it is likely that neighboring locations of certain points are empty without any features, which implies a waste of computation on meaningless locations. Therefore, we can generalize Eq.~\ref{eq:dense_convolution} with Eq.~\ref{eq:sparse_convolution}, which is defined as:
\begin{equation}
    \mathbf{x}^{\text{out}}_\mbf{u} = \sum_{\mbf{i} \in \mathcal{N}^D(\mbf{u}, \mathcal{C}^{\text{in}})} W_\mbf{i} \mathbf{x}^{\text{in}}_{\mbf{u} + \mbf{i}} \text{ for } \mbf{u} \in \mathcal{C}^{\text{out}}
\label{eq:sparse_convolution}
\end{equation}
where $\mathcal{N}^D$ is a set of offsets that define the shape of a kernel. $\mathcal{N}^D(\mbf{u}, \mathcal{C}^\text{in})= \{\mbf{i} | \mbf{u} + \mbf{i} \in \mathcal{C}^\text{in}, \mbf{i} \in \mathcal{N}^D \}$ is the set of offsets from the current center, $\mbf{u}$, which can be arbitrarily defined to describe the shape of the convolutional kernels. Notice that the computation is only carried between $\mathcal{C}^{\text{in}}$ and $\mathcal{C}^{\text{out}}$ that are non-empty locations.

During the implementation of the generalized sparse convolution, three steps are involved: 1) Generate the output coordinates $\mathcal{C}^{\text{out}}$ when the input coordinates $\mathcal{C}^\text{in}$, the convolution layer stride size, the input sparse tensor stride size are given~\cite{choy20194d}; 2) Establish the mapping relationship between the input and output coordinates for each kernel weight $W_\mbf{i}$ used to link the inputs, the kernel weights, and the outputs; 3) Conduct the computation of Eq. \ref{eq:sparse_convolution} by iterating the kernel weights over all corresponding input-to-output mappings and input features.

The extra overhead of Eq. \ref{eq:sparse_convolution}  compared to Eq. \ref{eq:dense_convolution} is the need of constructing and maintaining the mapping relationship in step 2. This involves extensive insertions as well as  search and can become the main bottleneck when the number of points is huge. In this paper, it is implemented using a GPU-based hashmap which reduces the overhead.

\textbf{Pipeline Parallelism with SparseDNN}. Existing parallel approaches can only support dense tensor computing, which inevitably results in high memory requirements. In SparsePipe, we instead aim at conducting sparse computations across multiple GPU processors while partitioning the DNN model to different GPU processors. The generalized convolution operations (Eq.~\ref{eq:sparse_convolution}) are computed to send the intermediate results immediately to the next stage during the forward computation while collecting results during the backward computation as described in Section \ref{subsec:pipeline}.

To meet the above requirements, efficient communication functions specified to sparse tensor are implemented, by adopting Pytorch Distributed Parallel Library\cite{li2020pytorch}. The inter-GPU communication is implemented with the Gloo distributed communication package by exchanging essential information for representing sparse tensor (namely the coordinate matrix and feature matrix). Similar to PipeDream \cite{narayanan2019pipedream}, round-robin scheduling is adopted to make sure gradients computed in the backward pass are routed to the corresponding processor from the forward pass, which guarantees consistent computing for a single round of forward-backward passes.

SparsePipe incorporates the latest Pytorch \cite{paszke2017automatic} and MinkowskiEngine \cite{choy20194d}. There exist many other alternative frameworks for conducting sparse computation, including, but not limited to, SparseConvNet \cite{graham20183d} and  SpConv\footnote{\url{https://github.com/traveller59/spconv}}. We are planning to release our SparsePipe implementation to facilitate the research in this direction.

\subsection{Partitioning for Heterogeneity} \label{sec:method_hete}
Like other pipelining approaches, the throughput of SparsePipe is determined by its slowest stage. Vastly different throughputs for stages will result in potential bubbles leading to load imbalance and resource under-utilization. Naive model partitions that are heuristically determined by the practitioners are often point solutions, which are difficult to obtain a balanced partition. The partitioning approach of PipeDream \cite{narayanan2019pipedream} assumes that the GPUs or servers used in the pipeline are homogeneous. This is generally not the case because clusters grow organically and consist of a wide variety of GPUs due to the short release cycle of new  GPU architectures.  Pipe-torch~\cite{zhan2019pipe} proposes a model partitioning algorithm suitable for different network bandwidths.

In the following, we will present a heterogeneous-aware pipeline model partition algorithm (Fig. \ref{fig:hpc}) that achieves load balance based on the heterogeneous GPUs, where ``heterogeneous” refers to the configuration of GPUs with different computational abilities.

\begin{figure}[hbt!]
    \centering
    \includegraphics[width=0.5\textwidth]{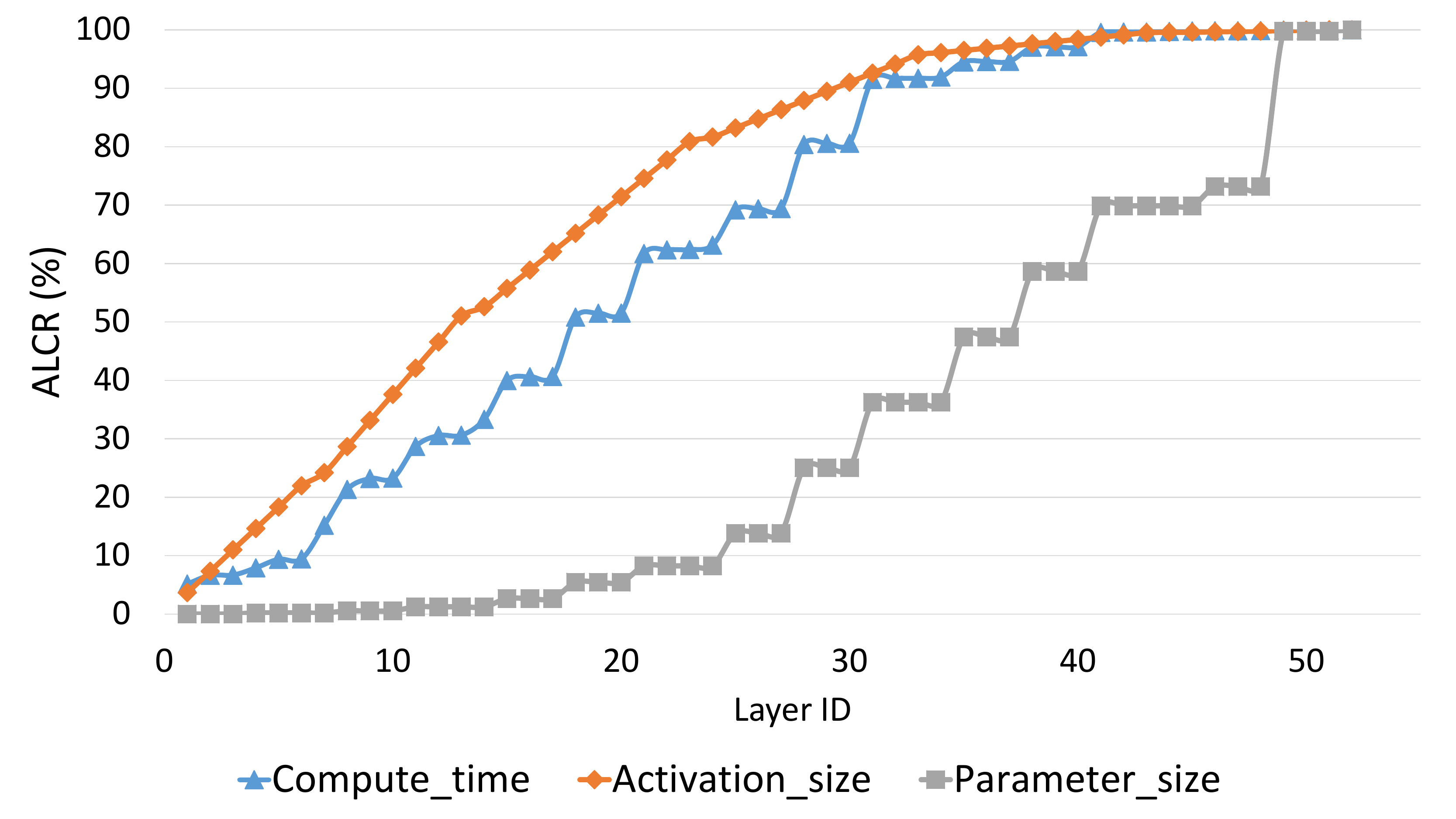}
    \caption{The accumulative layer cost ratio (ALCR) of compute time, activation size and parameter size for each layer of VGG16-BN model on Titan XP.}
    \label{fig:compute_time_cdf}
\end{figure}

\textbf{Workflow of SmartProfile}. Before partitioning the model, we need to derive the time incurred by each layer. SparsePipe assumes that the computational time during the model training remains the same for different runs in the same machine. Thus, a profiling script (\textit{SmartProfile})  is utilized to estimate each DNN layer’s computation and communication cost on different types of GPUs. This helps the partition algorithm generate a reasonable work partition and layer assignment. It begins with collecting a set of configurations such as batch size, GPU configurations, and model types, followed by a \textbf{warmup} stage to keep GPUs busy before the actual profiling. The warmup is done by running 50 iterations of DNN training. Another 100 iterations are added in the \textbf{profile} stage to record logs of each model layer and dump to text files, which consists of the total computation time (of both forward and backward computation), activation size, and weight parameter size.

For multiple GPU training, we will run this profiling independently on each GPU with a different computational ability. The partition algorithm will take all the records obtained from all types of GPUs, together with other information such as bandwidth, number of workers, and output a feasible model partitioning solution including 1) the layer-to-stage assignment and 2) the number of workers for each stage.

One visual example of the profiling is demonstrated in Fig. \ref{fig:compute_time_cdf}, where we accumulate the computational time, parameter sizes, and activation sizes over layer indexes of the VGG16~\cite{simonyan2014very} with batch normalization \cite{ioffe2015batch} (VGG16-BN) model on one Titan XP device. We observe that most of the computation comes from early layers with larger activation sizes whereas later layers have more weight parameters.

 \textbf{Overhead of SmartProfile}. While training deep learning models usually takes hours or days (e.g., roughly 8/3 hours of training for Dense/Sparse DNN in our experiments), the overhead of running the profiler is much smaller and it could finish in 5 minutes for one GPU type (about 1\% - 3\% of the training time). Multi-GPU training will repeat the profiling on each GPU type thus slightly increasing the overhead. Besides, this partitioning is only a one-time configuration for a model.

SparsePipe extends from the profiler in PipeDream regarding a different workload distribution. In PipeDream, they assume all the GPUs are of the same type. However, this is not always the case in reality. Our SparsePipe instead considers the heterogeneity that GPUs used during the training might have different computational abilities. SparsePipe can balancedly distribute the work among varying types of GPUs thus making the time spent by each GPU about the same. This minimizes the idling time and  achieves more effective load-balancing and better speedups than PipeDream. The workload partition and distribution of SparsePipe works as follows.

To load balance the  model partitioning, we need to divide it into stages such that each stage has similar execution time  (or  throughput) and the  communication overhead is minimized across stages. SparsePipe (just like PipeDream) allows replicating stages on multiple processors, therefore, being able to speed up the slowest stage of the pipeline. 
\begin{algorithm}[hbt!]
\caption{Hete-aware pipeline model partition algorithm}
\label{alg:partition}
\begin{algorithmic}[1]
    \State get the profiling results of each GPU with different computational time
    \State sort GPUs, and store as $gpu\_list$
    \State // initialize data parallel time as the baseline time
    \State obtain the data parallel time $dp\_comp$ from layer $i$ to layer $j$ ($i <= j$) with a range of GPUs in $gpu\_list$ by calling $GetCompTime$ in Algorithm \ref{alg:getCompTime}
    \State // Get the minimum pipeline partition time
    \State Initialize $min\_pipe$ as a 3D array with initial value from $dp\_comp$
    \For{i = 0, L}
         \For{j = i+1, L}
              \For{m = 0, length($gpu\_list$)}
                  \For{k = i, j}
                      \State get the minimum stage time from layer i to k with $gpu\_list[:m_1]$ from $min\_pipe[i][k][m_1]$, $m_1 \in [0, m]$
                      \State get the minimum stage time from layer k+1 to j with $gpu\_list[m_1:m]$ by calling $GetCompTime$ in Algorithm \ref{alg:getCompTime}
                      \State get the communication time between layer $k$ to $k+1$
                      \State get the maximum time of the above three numbers as the possible solutions to stage ranging from layer $i$ to $j$ with $m$ GPUs
                  \EndFor
                  \State find the minimum time $mini\_comp$ from the above possible solutions 
                  \If{$mini\_comp < min\_pipe[i][j][m]$}
                      \State update $mini\_pipe[i][j][m]$ and the split position $k$
                  \EndIf
              \EndFor
           \EndFor
    \EndFor
    \Return $min\_pipe$
\end{algorithmic}
\end{algorithm}
The overall model partition is now formulated: Given a DNN model of $L$ layers and a set of $M$ heterogeneous GPU processors, the goal is to partition the model and assign the partitions to GPUs, such that the total computational time in one iteration is minimized. Formally, let $\mathrm{S}(n)$ denote a set with a cardinality of $n$, whose elements come from GPU identified as $\{1,2,..,n\}$. Denote by $C(i,j,\mathrm{S}(n))$ the  time taken by the slowest stage in the pipeline from layer $i$ to layer $j$ using a processor set $\mathrm{S}(n)$. The goal of the algorithm is to find $C(0, L, \mathrm{S}(M))$, the corresponding partition stages, and GPU assignments.

To obtain the solution, let $Q(i, j, \mathrm{S}(m))$ represent the overall time taken by a single stage ranging from layer $i$ to $j$ replicated over  processor set $\mathrm{S}(m)$. $Q$ includes both the computation and parameter synchronization time and thereby can be computed as:
\begin{equation}
    Q(i,j,\mathrm{S}(m)) = \frac{1}{m}(\max_{a \in \mathrm{S}(m)} \sum_{l=i}^{j} t_a^l + \frac{2(m-1)\sum_{l=i}^{j}p^l}{BW}),
\label{eqn:dp}
\end{equation}
\noindent where $t_a^l$ refers to the computational time of layer $l$ on processor $a$, which belongs to set $\mathrm{S}(m)$. $p^l$ refers to the weight parameter size for layer $l$, $BW$ is the bandwidth among GPUs. The first term refers to the total computational time for all layers in this stage. Because of the heterogeneous GPUs, the computational time is determined by the slowest processor\eatme{ list from all possible} in set $\mathrm{S}(m)$. The second term stands for the communication time needed when synchronizing the weight parameters with the current stage, where we used an efficient all\_reduce collective communication \cite{narayanan2019pipedream}. 

\begin{algorithm}[bt!]
\caption{GetCompTime function called by Algorithm \ref{alg:partition}}
\label{alg:getCompTime}
\begin{algorithmic}[1]
    \Require Layer $i$, Layer $j$, GPU list $mList$
    \Ensure The data parallel time from layer $i$ to layer $j$ using machines $mList$
    \Function{GetCompTime}{$i$, $j$, $mList$}
        \State Get the slowest GPU among $mList$, named $mSlow$
        \State Get the total computational time from Layer $i$ to Layer $j$ on GPU $mSlow$, assign to $compSum$
        \State Get the total parameter size from $i$ to $j$, named $paraSum$
        \State $m$ = number of GPUs in $mList$
        \State Estimate the communication time of $paraSum$ among $m$ GPUs, assigned as $commTime$ 
        \Return $sum(compSum, commTime)/m$
    \EndFunction
\end{algorithmic}
\end{algorithm}

The problem of determining the minimum pipeline time, $C(i,j,\mathrm{S}(M))$, can now be divided into sub-problems consisting of a minimum sub-pipeline of the time from layer $i$ to $k$ using a processor set $\mathrm{S}(N)$, followed by a stage from layer $k+1$ to $j$ with the remaining $M-N$ processors, which we denote as $\mathrm{S}(M-N)$ for simplicity. This can be expressed as the following equation:
\begin{equation}
\begin{aligned}
    C(i, j, \mathrm{S}(M)) = &\min_{i \leq k < j} \min_{\mathrm{S}(N) \subset \mathrm{S}(M)} \max ( C(i, k, \mathrm{S}(N)), \\
    &\frac{a_k}{BW}, Q(k+1, j, \mathrm{S}(M-N))),
\end{aligned}
\label{eqn:partition}
\end{equation}
where $\frac{a_k}{BW}$ is the communication time between these two stages, with $a_k$ defining the activation size between layer $k$ and $k+1$ that is obtained during profiling. This problem can be solved using dynamic programming and backtracking.

Through recursion, the problem of obtaining the solution from layers $i$ to $j$ with GPU set $S(M)$ has been converted to a relatively smaller problem of obtaining two solutions from layer $i$ to layer $k$ with a subset of $S(M)$ and from layer $k+1$ to $j$ with the remaining GPUs in set $S(M)$, where $k$ is between $i$ and $j$. To figure out the optimal configuration, we need to partition the processor set into two subsets and try all the possible combinations, which is $O(2^n)$. A simple approach trying all options would potentially require non-polynomial time.  We use a simple but effective heuristic that is based on sorting the processors using computational capability. Our approach has two advantages: 1) The GPUs, that assigned to the same stage with consecutive layers, are generally homogeneous - this guarantees load balancing among GPUs thereby improving the computation efficiency, and 2) the number of choices is now limited to O(n). For example, with a GPU list of $[m_1, m_2, m_3,m_4]$, we only consider the possible solutions of ($[m_1]$, $[m_2, m_3, m_4]$), ($[m_1, m_2]$, $[m_3, m_4]$), and ($[m_1, m_2, m_3]$, $[m_4]$). That is what we did in our experiments. Our results in Fig.~\ref{fig:sparse_mp_vs_hetemp} demonstrate that this is an effective heuristic. We plan to investigate other heuristics for finding optimal partitioning as well.

The heterogeneous-aware pipeline model partitioning algorithm is shown in Algorithm~\ref{alg:partition}. Our algorithm begins with obtaining the forward and backward computational time with each layer on each GPU type by calling the profiling script. Then the GPU set $S(M)$ is sorted according to the heuristic strategy mentioned above, and this is the order of GPUs used during model partition. After that, the data-parallel computational time is computed, which serves as a baseline solution. At last, the problem of getting the pipeline time $C(i, j, S(M))$ is obtained by recursively calculating the sub-pipelines. Algorithm \ref{alg:getCompTime} shows how to get the total data-parallel time regarding the GPU list. In the data-parallel, the weight parameters need to be synchronized during every iteration; therefore, the computational time is determined by the last GPU finishing the computation.

 \textbf{Discussion:}
The main improvements of SparsePipe over PipeDream are the following: 1) We support heterogeneous platforms that consist of GPUs with different computational powers, 2) We extend to sparse computations while achieving lower memory overhead and better speed-accuracy tradeoffs. The latter is extremely important to 3D data processing.

\vspace{-0.2cm}
\section{Evaluation}\label{sec:experiment}
This section evaluates the effectiveness of our proposed SparsePipe framework on two server clusters. These results show that a combination of pipelined model parallelism and data parallelism is superior to using only data parallelism. The accuracy of SparsePipe machine learning models have comparable or better accuracy than the dense counterparts. Additionally, the SparsePipe framework can be effectively partitioned for a heterogeneous  GPU system. We outline the details in the following subsection.

\vspace{-0.2cm}
\subsection{Experimental Setup}\label{sec:exp_setup}
We have two available servers for the experiments. The first server is with one Titan V GPU and three Titan XP GPUs. The second one was equipped with four GeForce RTX 2080 Ti. The detailed GPU specification comparisons can be found in Table \ref{tab:gpu}. All servers run a 64-bit Ubuntu system with CUDA toolkit 10.2 and CuDNN v7.6.

To evaluate the efficiency of the proposed SparsePipe and heterogeneous-aware partition algorithm, we used the classic classification model VGG16\cite{simonyan2014very} with batch normalization \cite{ioffe2015batch}. We trained and evaluated the developed models on the ModelNet40 dataset \cite{wu20153d}, which contains 40 shape categories of CAD models. We measured the average epoch time for model training and its training and testing accuracy.

\textbf{Implementation Details.} For dense tensor representation, we generated voxel grids for the shapes by calling function  $TriangleMeshToVoxelGrid$ using Kaolin with efficient implementations for 3D data preprocessing \cite{kaolin2019arxiv}. For sparse tensor representation, we sampled point clouds from its triangle meshes with $TriangleMeshToPointCloud$. We sampled up to $16,384$ points for each sample and cached them for a faster I/O. Then $4,096$ points are randomly selected from them. For all the experiments, we trained the models for 90 epochs using the SGD optimizer with a learning rate of $10^{-2}$, and set the momentum to 0.9.

\begin{table}
\centering
\caption{GPU Configurations of our servers.} \label{tab:gpu}
\begin{tabular}{|c|c|c|c|}
\hline
Server                                                                  & \multicolumn{2}{c|}{1st}                    & 2nd                         \\ \hline
GPU                                                                     & TITAN V $ \times 1$ & TITAN XP $\times 3$ &  RTX 2080 Ti $ \times 4$ \\ \hline
Architecture                                                            & Volta               & Pascal              & Turing                     \\ \hline
\begin{tabular}[c]{@{}c@{}}CUDA \\ Cores\end{tabular}                   & 5120                & 3840                & 4352                       \\ \hline
\begin{tabular}[c]{@{}c@{}}Boost \\ Clock (MHz)\end{tabular}                                                        & 1455               & 1770              & 1545                     \\ \hline
\begin{tabular}[c]{@{}c@{}}Single \\ Precision \\ (TFLOPS)\end{tabular} & 13.8                & 12.1                & 13.4                       \\ \hline
\begin{tabular}[c]{@{}c@{}}Memory\\ Size (GB)\end{tabular}              & 12                  & 12                  & 11                         \\ \hline
\begin{tabular}[c]{@{}c@{}}Memory BW\\ (GB/sec)\end{tabular}            & 653                 & 547                 & 352                        \\ \hline
\end{tabular}
\vspace{-0.5cm}
\end{table}

\begin{table*}[hbt!]
    \centering
    \caption{Performance evaluation with different random input dropouts for Sparse DNN at resolutions of $32 \times 32 \times 32$ and $50 \times 50 \times 50$. All modoels are evaluated and trained using single GPU or 4 GPUs with data parallelism. The speed-ups are computed against a base model -- the Dense DNN  trained on single GPU -- the D-1 (ref.). ``D/S-p" denotes Dense/Sparse DNN with a certain point sampling ratio $p\in\{0.25,0.5,1\}$. For Dense DNN, the convolution computation is conducted by densely iterating every voxel in the 3D space. Instead, Sparse  DNN  conducts  an  efficient  computation  only  on  non-empty voxels thus avoiding unnecessary computation on empty locations. The main conclusions are: Compared to Dense DNN, Sparse DNN 1) has a faster training speed, 2) can further increase the training speed by exploiting more point sparsity, and 3) is memory efficient that allows to feed the inputs of a large batch size.
    } \label{tab:sparsity}
\begin{tabular}{c|c|c|c|c|c|c|c|c|c}
\hline  
        &      &    \multicolumn{4}{c|}{1 GPU}  & \multicolumn{4}{c}{4 GPUs} \\ \hline
Resolution & D/S-p & D-1 (ref.) &  S-1 &  S-0.5 & S-0.25   & D-1 & S-1 &  S-0.5 & S-0.25  \\ \hline \hline
\multirow{5}{*}{\makecell{$32 \times 32 \times 32$ \\ (Voxel Occupancy Ratio 7.6\%)}} 
                  & Training Acc (\%) &  99.32 &  99.95  &  99.81 &  99.72 & 99.46 & 99.94      & 99.70  &  99.19    \\ \cline{2-10}
                  & Testing Acc (\%) &  84.46 &  86.28 &   85.39 &  84.55  & 84.19 &  85.29    & 84.04 &  84.12    \\ \cline{2-10}
                 & Epoch Time (sec.) &  82.71 &   62.04 &  47.43 &  36.66  & 24.45 &  17.54 &   14.16  &    11.14  \\ \cline{2-10}
                 & Speedup  &   1 &  1.33 &   1.74 &  2.26 &   3.38 &  4.72   &   5.84   &     7.42   \\ \cline{2-10}
                 & Batch Size  & 128  &  256  &  256 &  512  & 128 &  256    &  256  &  512  \\ \cline{2-10}
\hline \hline

\multirow{5}{*}{\makecell{$50 \times 50 \times 50$\\ (Voxel Occupancy Ratio 2.6\%) }}
        &  Training Acc (\%)  &   99.34 &  99.88 &   99.82 &   99.63   & 99.36 & 99.93 &  99.66  &  99.25      \\ \cline{2-10}
       &   Testing Acc (\%)  &   86.78 &   87.61 &    87.41 &   86.93 &  87.18 & 87.80 &   87.52&  87.00    \\ \cline{2-10}
        &   Epoch Time (sec.) &   314.94  &   129.63  &   99.70  &   71.43  & 82.05 &   35.11   &   27.62 &    20.37 \\ \cline{2-10}
        &   Speedup  &   1 &  2.43  & 3.16 &  4.41  &   3.84 &     8.97   &    11.40   &     15.46  \\ \cline{2-10}
        &   Batch Size  &   32 &  128  &  128 &  256  & 32  & 128    &  128  &  256   \\ \hline
\end{tabular}
\vspace{-0.5cm}
\end{table*}
\subsection{Comparison to Dense 3D Computation} \label{sec:exp_sparse_vs_dense}

\eatme{
We compared SparsePipe (Sparse DNN) against its dense 3D convolution version (referred to as Dense CNN). In Dense CNN, the point cloud space was quantized into voxels, and each voxel has a binary state, occupied or unoccupied by point clouds. A fixed occupancy grid of size $32 \times 32 \times 32$ was chosen for the voxels. We could not execut a higher resolution for Dense DNN  since it significantly increases the memory consumption and causes an out-of-the-memory issue. Table \ref{tab:dense_vs_sparse} shows the experimental results on ModelNet40 dataset with Dense CNN. The training and testing accuracy are listed. To make a fair comparison, in SparsePipe, we quantized the sampled points and divided the spanned space into the same voxel resolution of $32 \times 32 \times 32$ as the dense CNN did. Compared to the dense 3D CNN, SparsePipe achieves higher performance in terms of accuracy with higher resolution. 

To validate that a higher resolution could help in discriminating shapes by providing finer details, we gradually increased the resolutions to $50\times50\times50$ and $100\times100\times100$. All the models were trained with a single GPU to make a direct comparison. The results are presented in Table \ref{tab:dense_vs_sparse}. Both $50\times50\times50$ and $100\times100\times100$ are comparable and achieve a higher accuracy. $50\times50\times50$  obtains a better speed and accuracy trade-off. }

We compared SparsePipe (Sparse DNN) against its dense 3D convolution version (referred to as Dense DNN). In Dense DNN, the point cloud space was quantized into voxels, and each voxel has a binary state, occupied or unoccupied by point clouds. A fixed occupancy grid of size $32 \times 32 \times 32$ and $50 \times 50 \times 50$ were chosen for the voxels. For the grid size of $32 \times 32 \times 32$, the average voxel occupancy ratio, which is calculated as the ratio between the number of occupied voxels and the total number of voxels in a point cloud, is around 7.6\%. The voxel occupancy ratio for $50 \times 50 \times 50$ is around 2.6\%. We could not effectively execute  higher resolution for Dense DNN since it significantly increases the memory consumption and causes an out-of-the-memory issue. 

\begin{figure}[bt!]
    \centering
    \includegraphics[width=0.48\textwidth]{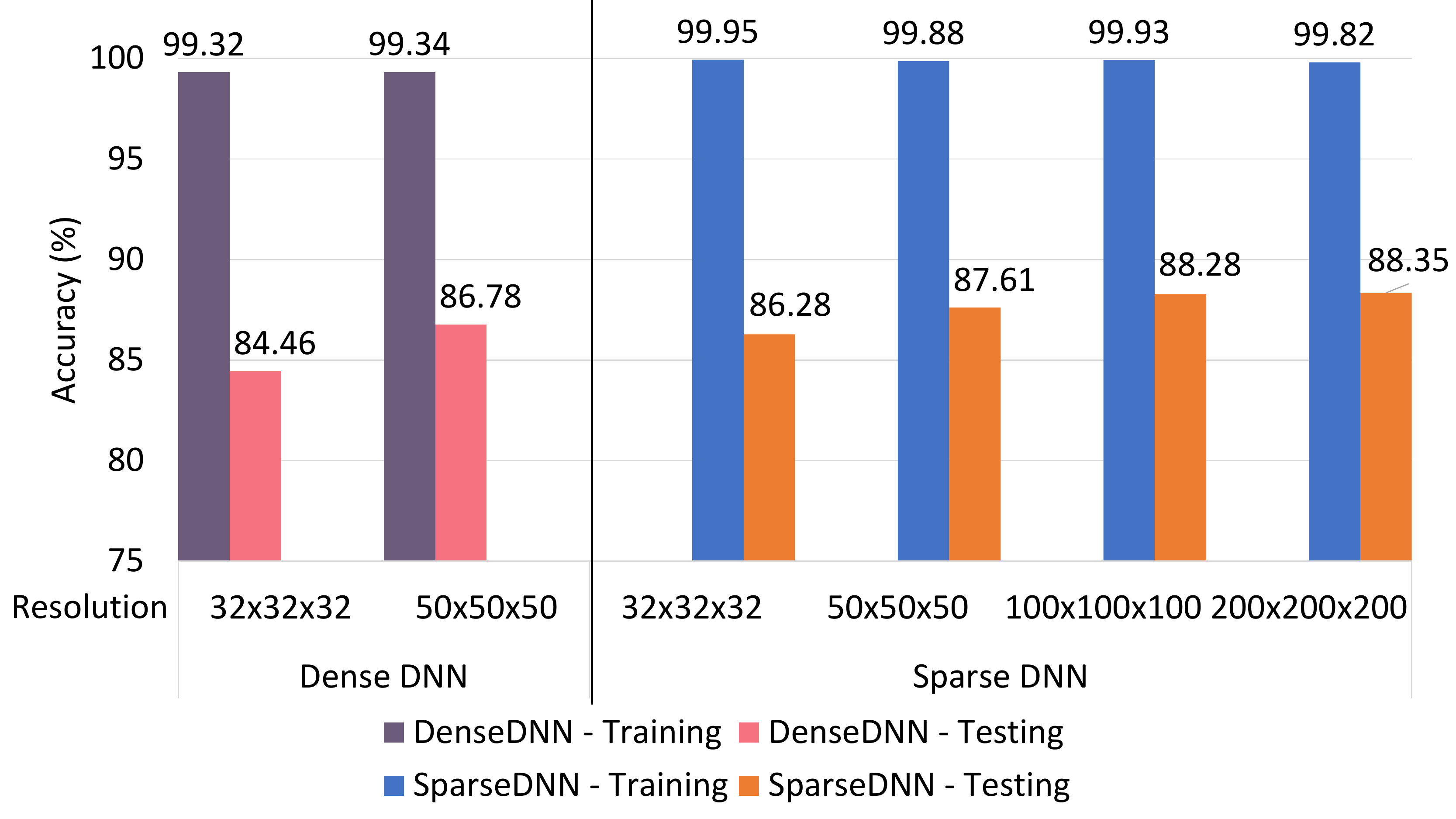}
    \caption{Compared to Dense DNN, Sparse DNN is significantly less memory intensive. Sparse DNN can support a resolution of $200 \times 200 \times 200$ while   Dense DNN is limited to $50 \times 50 \times 50$. This opens up the possibility of solving problems at larger scale. Both models can achieve higher accuracy with a higher resolution.}
    \label{fig:dense_vs_sparse}
\vspace{-0.7cm}
\end{figure}

\textbf{Memory Efficiency with Sparse DNN.} Fig. \ref{fig:dense_vs_sparse} shows the experimental results on ModelNet40 dataset with Sparse DNN and Dense DNN using different resolutions. The training and testing accuracy are presented. To make a fair comparison, in SparsePipe, we quantized the sampled points and divided the spanned space into the same voxel resolution of the dense DNN.  We observe that either for Dense DNN or Sparse DNN, a higher voxel resolution could help increasing the testing accuracy. This is intuitive because finer resolution can provide more details in discriminating shapes. Notice that Dense DNN can not conduct a higher resolution of $100 \times 100 \times 100$ or $200 \times 200 \times 200$ because of the large memory footprint. This shows that Sparse DNN is memory friendly, allowing larger input resolutions to achieve higher accuracy. Though at the same resolution Sparse DNN achieves higher performance in terms of accuracy compared to the Dense DNN, this variation in accuracy is partly due to the quantization and sampling as described in ''Implementation Details'' Section \ref{sec:exp_setup}.

\textbf{Speed-ups with Point Sparsity.} We explored and exploited the point sparsity for accelerating the training of Sparse DNN. Unlike Dense DNN where the computation is permanently fixed once the model layers and resolutions are determined, Sparse DNN can further achieve a faster training speed by dropping a subset of points while not sacrificing too much accuracy. To validate it, we introduced a dropout ratio $\theta$ that uniformly sampled from $[0, p] $ where $p \leq 1$. Under this concept, $p=0$ means dropping all the points whereas $p=1$ keeps all points. We evaluated the model across varied uniformity $p\in\{0.25, 0.5, 1\}$)) as shown in Table \ref{tab:sparsity} with two different resolution experiments conducted, i.e., $32 \times 32 \times 32$ and $50 \times 50 \times 50$. This dropout ratio of $\theta$ will end up with a reduced number of input points for training and testing the model. For Dense DNN, since the point sparsity doesn't influence the computation or memory given a fixed resolution, the result of $p=1$ is shown in Table \ref{tab:sparsity}. These experiments are conducted on 1 GPU or 4 GPUs with data parallelism. Please note that, ``data parallelelism” here refers to the naive parallelism where the entire neural network replicates among multiple GPUs. For SparsePipe, data parallel works the same way as Dense DNN. It can be viewed as the DNN partitioned into one stage replicated among multiple GPUs.

\begin{table*}[hbt!]
\centering
\caption{Summary of GPU configurations, model split and stage assignment details of comparing HETE-MP with MP. A split config of ``4-1” means the model is split into 2 stages where the first stage is replicated among 4 GPUs and the second stage on one GPU. The ``Stage Assignment” lists the assigned GPUs for each stage.} \label{tab:sparse_mp_vs_hetemp}
\begin{tabular}{|c|c|c|c|c|c|c|}
\hline
\#GPUs & GPU Config & MP/HETE-MP & Split Config & Stage Assignment \\  \hline  \hline
5      &  1 TitanV + 4 RTX  & MP    &   1-2-1-1  & -  \\
5      &  1 TitanV + 4 RTX  & HETE-MP&      4-1  & (4 RTX) (1 TitanV)   \\ \hline  \hline
6      & 1 TitanV + 1 TitanXP + 4 RTX & MP & 4-1-1  & -  \\
6      & 1 TitanV + 1 TitanXP + 4 RTX & HETE-MP & 4-1-1 & (4 RTX) (1 TitanV) (1 TitanXP)  \\ \hline \hline
7      & 1 TitanV + 2 TitanXP + 4 RTX & MP & 4-1-1-1  & -  \\
7      & 1 TitanV + 2 TitanXP + 4 RTX & HETE-MP & 6-1  & (4 RTX + 2 TitanXP) (1 TitanV) \\ \hline \hline
8      & Server 1 + Server 2      &  MP  &  4-1-1-1-1   & - \\ 
8      &    Server 1 + Server 2 & HETE-MP & 7-1  & (4 RTX + 3 TitanXP) (1 TitanV)\\ \hline
\end{tabular}
\vspace{-0.5cm}
\end{table*}

\eatme{In Table \ref{tab:sparsity}, sparse {\color{blue}DNN} maintains relatively stable accuracy while it is significantly faster (e.g., $5.04 \times$ faster with $p = 0.25$) with fewer points. SparsePipe is memory friendly compared to Dense DNN which is equipped with the latest CUDA and CuDNN. SparsePipe allows a larger batch size during training, up to 1,024 compared to 128 for Dense DNN. The current implementation of SparsePipe is faster than Dense DNN while achieving comparable performance with $p \leq 0.25$. A more efficient SparsePipe can loosen the condition to achieve a target acceleration with denser point sampling (e.g., $p < \leq 0.3$).
It is promising to speed up our current solution in a future version given the availability of many advanced techniques such as more coalesced memory access via block-based sparse convolutions \cite{ren2018sbnet} or more efficient GPU hashtables for fast indexing.}

In Table \ref{tab:sparsity}, the experiment conducting with Dense or Sparse DNN using dropout probability $p$ is abbreviated as $D/S-p$. For Dense DNN, the convolution computation is conducted by densely iterating every voxel in the 3D space. Instead, Sparse DNN conducts an efficient computation only on non-empty voxels thus avoiding unnecessary computation on empty locations. Compared to the baseline model $D-1$, we achieve a speedup of $1.33$ at the resolution of $32\times 32\times 32$ and is $4.72\times$ faster when training models with 4 GPUs. We also demonstrate that: By dropping more points, our model remains relatively stable on the accuracy while further achieving a larger speedup (up to $7.42\times$ faster compared to the model $D-1$ with $p = 0.25$).

We further choose to increase the input resolution of the point cloud from  $32\times 32\times 32$ to $50 \times 50 \times 50$, which allows us to achieve a better accuracy. At a higher resolution, the baseline $D-1$ incurs a larger overhead on the computation, while Sparse DNN $S-1$ is faster demonstrated by a speedup of $2.43$ with single GPU training, and by that of $8.97$ with the 4-GPU training. Sparse DNN further achieves a $15.46\times$ speedup on 4 GPUs with point sparsity $p=0.25$.

The memory requirements of SparsePipe (for the given sparsity level of typical datasets) is much smaller compared to Dense DNN that leverages  the latest CUDA and CuDNN. The batch size shown in Table \ref{tab:sparsity} is the largest one that can fit into a GPU memory either for dense or sparse model. As the table shows, SparsePipe allows a larger batch size during training, up to 512 compared to 128 for Dense DNN at the resolution of $32 \times 32 \times 32$, and up to 256 compared to 32 for Dense DNN at the resolution of $50 \times 50 \times 50$. In summary, Table \ref{tab:sparsity} shows that with Sparse DNN and point sparsity, the training process can be accelerated significantly and the memory storage can be efficiently utilized thus enabling higher resolution inputs to achieve higher accuracy. We can further speed up our current solution  leveraging  advanced techniques such as coalesced memory access via block-based sparse convolutions \cite{ren2018sbnet}.

\begin{figure}[hbt!]
    \centering
    \includegraphics[width=0.48\textwidth]{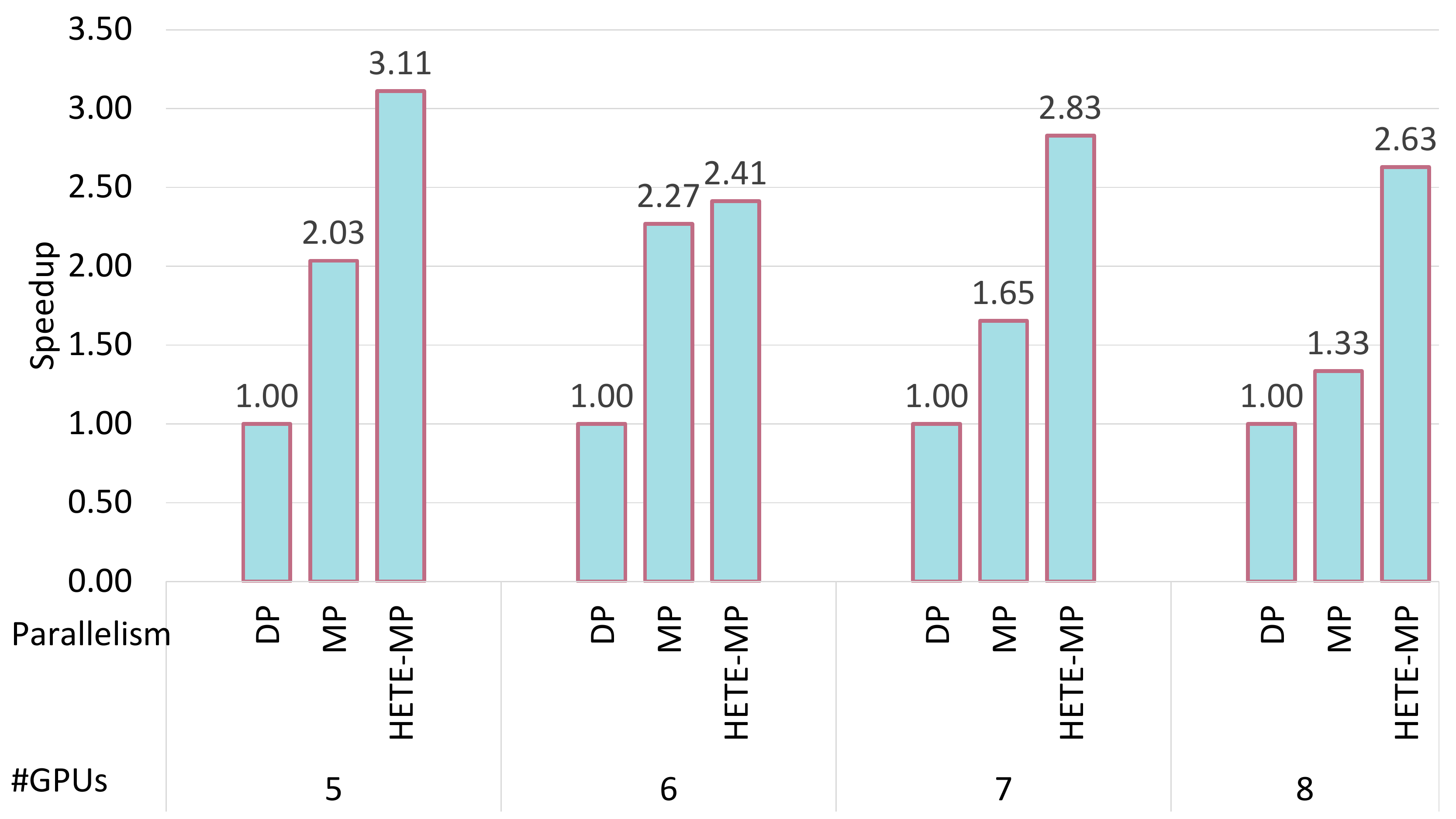}
    \caption{Performance evaluation between naive data parallelism (DP), pipeline model parallelism (MP) and heterogeneous aware pipeline model partition (HETE-MP). The speedups of MP and HETE-MP compared to DP are presented. The evaluation is conducted with two servers. All models are trained at the voxel resolution of size $50 \times 50 \times 50$ and a batch size of 64.}
    \label{fig:sparse_mp_vs_hetemp}
\vspace{-0.5cm}
\end{figure}

\vspace{-0.2cm}
\subsection{Comparison of Parallel Training Strategies} \label{sec:exp_parallel}
The advantages of SparsePipe over the dense convolution have been demonstrated in the previous section. We now explore the impact of data parallelism (DP) with pipeline model parallelism  and heterogeneous aware pipeline model partition (HETE-MP) on different number of GPUs across two servers. The results are presented in Fig. \ref{fig:sparse_mp_vs_hetemp}. We take the training time of the data parallel training (DP) as the baseline, and compute the speedups of MP and HETE-MP. The voxel resolution of $50\times50\times50$ is fixed. The batch size is $64$. 

The training time per epoch has been reduced a lot in the MP training, as shown in Fig. \ref{fig:sparse_mp_vs_hetemp}. Specifically, the speedups obtained are consistently larger and vary from 1.33 (on eight GPUs) to 2.27 (on six GPUs).  SparsePipe with pipeline model parallelism reduced the communication overhead via partitioning the model into several stages with some stages replicated among multiple GPUs and the last stage on $1$ GPU. The benefit is that large fully-connected layers were not replicated, thereby reducing communication overhead. Additionally, all processors are busy with the pipelining process, avoiding the GPU under-utilization.

We further demonstrated that HETE-MP can give better partition of models by exploiting different GPU characteristics, compared to MP which is with the assumption of homogeneous GPU processors across servers. To run these partitioning algorithm, the bandwidth between two servers needs to be specified in advance, which is measured with the iperf3 tool. Compared to MP that gets a speedups of $1.33-2.27$ over DP, HETE-MP can further accelerate the training and achieve higher speedups ranging from $2.41$ to $3.11$. For example, for five GPUs with one Titan V and four RTX, the speedup of HETE-MP is $3.11$. Comapred to MP achieving $2.03\times$ speedup, HETE-MP is $1.53\times$ faster. Similarly, on eight GPUs, HETE-MP can obtain $1.98 \times$ speedup compared to MP with a more load-balanced model partition.

We have shown that HETE-MP can run faster than MP by considering the difference of GPU computational abilities. Table \ref{tab:sparse_mp_vs_hetemp} lists the difference of workload partition and assignments between MP and HETE-MP on five to eight GPUs across two servers in details. A split config of ``4-1” means the model is split into 2 stages with the first stage replicated among 4 GPUs and the second stage on one GPU. The ``Stage Assignment” lists the assigned GPUs for each stage.

As presented in Table \ref{tab:sparse_mp_vs_hetemp}, MP and HETE-MP could generate different model split configurations. MP tends to split the model into multiple stages while balancing the time spent on each stage based on the layer costs obtained by SmartProfile on certain GPU. HETE-MP partitions the model in a more reasonable way by further taking the consideration of each GPU's computation ability. For example, when the number of GPUs is eight, MP partitions the model into five stages, with the first stage replicated among four GPUs with layers 0-23 and the rest stages with layers 24-52. For HETE-MP, after considering the difference of computational ability of GPUs, the model is partitioned into two stages with the first stage replicated among seven GPUs (four RTXs and three TitanXPs) with layers 0-23 and the second stage with layers 24-52 assigned to Titan V. The first stage takes most of the computational load in this model as shown in Fig. \ref{fig:compute_time_cdf} profiled on Titan XP with the second stage taking the rest of the computational load. With the latter assigned to Titan V, it would reduce the time spend on the second stage since Titan V has the highest computational ability among the three GPU types. Thus, the partition generated by HETE-MP can make full usage of each GPU's computational ability  and balance the workload among GPUs as much as possible. This demonstrates the benefits of taking heterogeneity into consideration, which balances the load among processors and guarantees the maximum of resource utilization.

\textbf{Discussions:} Pipeline model parallelism outperforms data parallelism, which is mainly due to the reduction of the communication to synchronize the network parameter. Thus, the pipeline model parallelism is more beneficial to models with large parameter sizes (e.g, VGGNet). For models with less parameters (e.g., ResNet), data parallelism is preferred. Besides, models with branching architecture such as 3D-UNet used for segmentation should avoid the current pipeline model parallelism due to too much intermediate results (even greater than the network parameters) needed to be communicated among GPUs thus increasing the communication time. It is promising to investigate the combination of branch neural networks and pipeline model parallelism in the future.

\vspace{-0.2cm}
\section{Conclusions} \label{sec:conclusion}
In this paper, we presented SparsePipe, an efficient and asynchronous parallelism approach for handling 3D point clouds that has a relatively small memory footprint as compared to dense approaches. It utilizes generalized convolutions with sparse tensor representation to build expressive high-dimensional convolutional neural networks. SparsePipe has reduced the communication overheads by integrating model parallelism with data parallelism and processes mini-batches in a pipelined fashion. Our heterogeneous-aware model partitioning algorithm can automatically partition the model layers among different processors and keeps the load as balanced as possible. The experimental results have demonstrated that SparsePipe obtained better performance on point cloud benchmarks compared to the dense counterpart. In the meantime, it achieved a larger training speed-up across multiple computing nodes when compared to the data parallelism.

\vspace{-0.2cm}
\section*{Acknowledgment} \label{sec:ack}
This work was funded by the U.S. Department of Energy, National Nuclear Security Administration, Advanced Simulation and Computing Program, as a Cooperative Agreement under the Predictive Science Academic Alliance Program, Contract No. DOE-NA0002378. This work was funded, in part, by the National Science Foundation, under Contract No. 1748652.

\vspace{-0.2cm}
\bibliographystyle{IEEEtran}
\bibliography{ref}

\end{document}